\documentclass[letterpaper]{article} 
\usepackage{aaai23}  
\usepackage{times}  
\usepackage{helvet}  
\usepackage{courier}  
\usepackage[hyphens]{url}  
\usepackage{graphicx} 
\usepackage{natbib}  
\usepackage{caption} 
\usepackage{algorithm}
\usepackage{algorithmic}
\usepackage{multirow}
\usepackage{amsmath}
\usepackage{amsthm}
\usepackage{mathrsfs}
\usepackage{amssymb}
\usepackage{booktabs}
\usepackage{rotating}
\usepackage{subfig}
\usepackage{booktabs}
\usepackage{bbding}
\graphicspath{{images/}}
\usepackage{newfloat}
\usepackage{listings}
\DeclareCaptionStyle{ruled}{labelfont=normalfont,labelsep=colon,strut=off} 
\floatstyle{ruled}
\newfloat{listing}{tb}{lst}{}
\floatname{listing}{Listing}
\title{DICNet: Deep Instance-Level Contrastive Network for Double Incomplete Multi-View Multi-Label Classification}
\author {
        Chengliang Liu\textsuperscript{\rm 1},
        Jie Wen\textsuperscript{\rm 1$\ast$},
        Xiaoling Luo\textsuperscript{\rm 1},
        Chao Huang\textsuperscript{\rm 2},
        Zhihao Wu\textsuperscript{\rm 1},
        Yong Xu\textsuperscript{\rm 1,\rm 3}\thanks{Corresponding authors.}
}
\affiliations {
    \textsuperscript{\rm 1} Shenzhen Key Laboratory of Visual Object Detection and Recognition, Harbin Institute of Technology, Shenzhen, China\\
    \textsuperscript{\rm 2} School of Cyber Science and Technology,
    Shenzhen Campus of Sun Yat-sen University, Shenzhen, China.\\
    \textsuperscript{\rm 3} Pengcheng Laboratory, Shenzhen, China.\\

    liucl1996@163.com, jiewen\_pr@126.com, xiaolingluoo@outlook.com, huangchao\_08@126.com, horatio\_ng@163.com, yongxu@ymail.com 
     
}

\usepackage{bibentry}

\begin{document}

\maketitle

\begin{abstract}
In recent years, multi-view multi-label learning has aroused extensive research enthusiasm. However, multi-view multi-label data in the real world is commonly incomplete due to the uncertain factors of data collection and manual annotation, which means that not only multi-view features are often missing, and label completeness is also difficult to be satisfied. To deal with the double incomplete multi-view multi-label classification problem, we propose a deep instance-level contrastive network, namely DICNet. Different from conventional methods, our DICNet focuses on leveraging deep neural network to exploit the high-level semantic representations of samples rather than shallow-level features. First, we utilize the stacked autoencoders to build an end-to-end multi-view feature extraction framework to learn the view-specific representations of samples. Furthermore, in order to improve the consensus representation ability, we introduce an incomplete instance-level contrastive learning scheme to guide the encoders to better extract the consensus information of multiple views and use a multi-view weighted fusion module to enhance the discrimination of semantic features. Overall, our DICNet is adept in capturing consistent discriminative representations of multi-view multi-label data and avoiding the negative effects of missing views and missing labels. Extensive experiments performed on five datasets validate that our method outperforms other state-of-the-art methods.
\end{abstract}

\section{Introduction}
As one of the important tasks of multi-label learning, the purpose of multi-label classification task is to label the observed samples with various category tags \cite{zhu2018multi,herrera2016multilabel}. For example, in the field of image recognition, a natural picture can be labeled with multiple labels such as `wild', `bird', and `sky'. Or in a text classification task, a piece of text can be classified into different semantic sets such as `soccer', `news', `advertising', etc. The broad application prospect of multi-label classification has aroused great research enthusiasm in both industry and academia \cite{liu2015low}. In addition, with the explosive growth of data sources and feature extraction methods, only describing, analyzing, and processing samples from a single perspective can no longer meet the needs of more complex and comprehensive analysis \cite{hu2020dmib,yuan2021adaptive,fang2021animc}. Doubtlessly, multi-view data collected from multiple sources is able to describe the observed objects more integrally and accurately \cite{li2021incomplete,li2020bipartite,wang2022mvsnet}. For example, in clinical practice, multiple indicators such as height, weight, and average hemoglobin are often used to synthetically diagnose whether a person is malnourished \cite{hu2019doubly,luo2021mvdrnet,ZhuMa2022}. Obviously, multi-view data is rich in more semantic information, which greatly facilitates the learning of multi-label semantic content \cite{huang2015learning,wang2020multi,hu2021view}. Therefore, different from the simple single-label classification task \cite{zhao2022shared}, this paper focuses on the multi-view multi-label classification task, namely MVMLC.

For MVMLC, a few meaningful methods have been proposed in recent years. Zhu et al. proposed a multi-view label embedding model, which learns the intermediate latent space through the Hilbert-Schmidt Independence Criterion \cite{gretton2005measuring} to indirectly bridge the feature space and the label space \cite{zhu2018multi}. Another representative matrix factorization (MF) based MVMLC method, named latent semantic aware multi-view multi-label learning (LSA-MML), aligns the semantic spaces by maximizing the dependencies of basis matrices corresponding to different views in the kernel space \cite{zhang2018latent}. In addition, deep neural networks (DNN) have also been developed to handle this issue \cite{liu2023incomplete}. For example, Fang et al. proposed the simultaneously combining multi-view multi-label learning (SIMM) neural network framework, which exploits adversarial loss and label loss to learn shared semantics and imposes the regularization constraint to obtain view-specific information \cite{fang2012simultaneously}. It is worth noting that these methods are invariably based on the unreasonable assumption that all views and labels are available. However, in practice, the data used for MVMLC is often incomplete. We consider this incompleteness from two aspects: On the one hand, the heterogenous data collected from multiple sources may be with missing views due to various reasons. For example, the media forms of records stored in archive may include text, audio, video, etc \cite{chen2022low,chen2022efficient}. These diverse media of information regarded as different views are not ubiquitously present in all records, so the multi-view features extracted from them are naturally incomplete; On the other hand, since it is difficult and expensive to manually tag all labels, label information belonging to real datasets is often missing to varying degrees, especially for observed objects with numerous strongly correlated labels \cite{fang2012simultaneously,zhang2018latent}. To sum up, unlike most existing works designed for the single-missing case, we contribute to dealing with the double-missing case, where random view-missing and label-missing occur simultaneously, \textit{i.e.}, double incomplete multi-view multi-label classification issue (DIMVMLC).

Obviously, both missing views and missing labels have serious impacts on multi-view multi-label learning \cite{xu2015multi,liu2020efficient,liu2022incomplete}. From the perspective of multi-view, missing views not only weaken the rich semantics of the original multi-view information, but also make the information fusion of unaligned multi-view more difficult due to the uncertain missing distribution compared with intact and aligned multi-view data. From the perspective of multi-label, the absence of non-specific labels not only impairs its supervision, but also poses challenges to the construction of unified learning model \cite{tan2017semi,huang2019improving,zhao2015semi}. Even so, some methods for incomplete multi-view learning (IMVL) and incomplete multi-label learning (IMLL) have been gradually proposed over the last few years, such as iMSF \cite{yuan2012multi}, iMvWL \cite{tan2018incomplete}, NAIML \cite{li2020concise}, etc (more detailed introduction in next section). Although these conventional methods represented by iMvWL have achieved certain results in the fields of IMVL and IMLL, it is the learning mode, which requires hand-designed feature extraction rules and is hard to generalize, that restricts the further development of incomplete multi-view multi-label learning. With the widespread popularity of deep learning, DNNs are increasingly applied in feature extraction and data analysis domain \cite{huang2022pixel,huang2022weakly}. Compared with conventional multi-view learning methods, DNN shows irreplaceable natural advantages \cite{wen2020dimc,li2022emocaps}.
Specifically, for one thing, traditional methods, whether based on MF, spectral clustering or kernel learning, are only capable of exploiting the shallow features of data \cite{zong2018multi,wang2019spectral,liu2019multiple}. However, capturing relatively high-level semantic content, which is DNN-friendly, is increasingly proving to be necessary, especially in complex multi-label classification tasks \cite{wen2020dimc}; for another, the performance of traditional multi-view learning models is heavily dependent on the parameter settings, and usually requires searching for optimal parameter combinations for different datasets. On the contrary, DNN enjoys the advantages of parameter adaptive learning, and the model design is more concise. In addition, the trained DNN has end-to-end reasoning capability, which makes it more suitable for application scenarios that require fast prediction instead of re-learning classification in database like most of traditional methods \cite{luo2023deep}. 

Therefore, in this paper, we propose a method named deep instance-level contrastive network (DICNet) for DIMVMLC. The DICNet consists of four parts: view-specific representation learning framework, instance-level contrastive learning module, weighted fusion module, and incomplete multi-label classification module. Inspired by \cite{zhang2019ae2}, our view-specific representation learning framework builds the basic high-level feature extraction and reconstruction network. On this basis, considering that instances corresponding to the same sample but from different views should contain consistent semantic features (i.e. consensus assumption), we introduce an incomplete instance-level contrastive loss on the extracted view-specific high-level representations to improve the cross-view consensus. Furthermore, in order to leverage the complementarity across views to obtain more discriminative semantic features, we design a weighted representations fusion module with prior missing-view information, which greatly mitigates the impact of the absence. Finally, a missing-label indicator matrix is introduced into the classifier to avoid invalid supervision information. Overall, compared with existing approaches, our DICNet proposed in this paper has the following outstanding contributions:
\begin{itemize}
\item To the best of our knowledge, our DICNet is the first DNN framework for the DIMVMLC task, which can cope with all sorts of incomplete cases, including missing labels and missing views. Furthermore, as a flexible end-to-end neural network, our approach is capable of training in a supervised or semi-supervised manner and performing real-time predictions, which is not possible with conventional methods.

\item Our DICNet focuses on extracting and learning high-level semantic features. On the one hand, the powerful incomplete instance-level contrastive learning guides the encoders to extract cross-view semantic features with better consensus. On the other hand, the weighted fusion strategy fully integrates complementary information without negative effects of missing views.

\item We conduct extensive experiments on five datasets, and the results establish that our DICNet significantly outperforms other benchmark methods on four key metrics.
\end{itemize}

\begin{figure*}[h]
	\centering
	\includegraphics[width=0.9\textwidth,height=8cm]{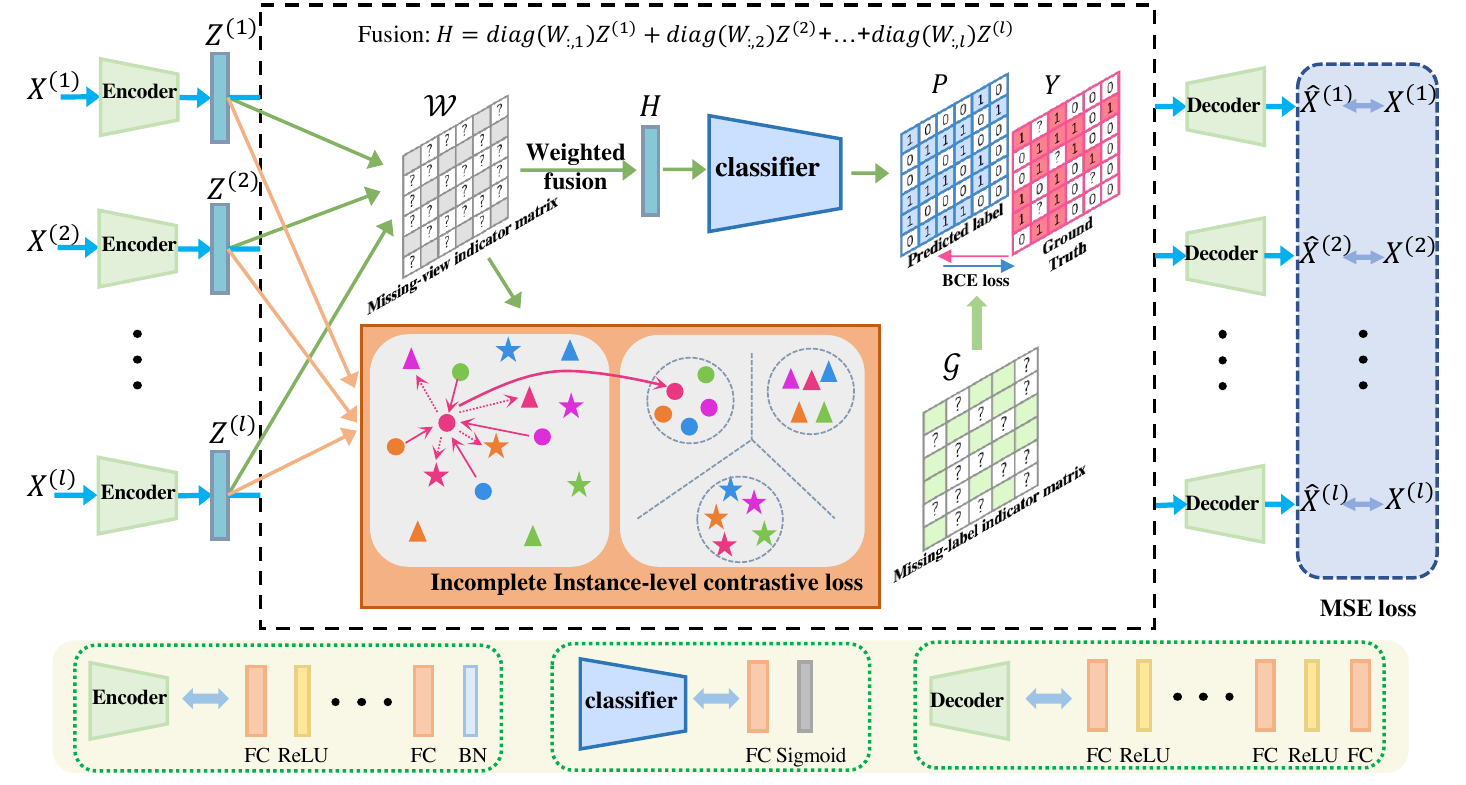}
	\caption{The main framework of our DICNet. The input data is processed by the encoders and then output to the weighted fusion module, the incomplete instance-level contrastive module, and the decoders, respectively. The structures of the encoders, decoders and classifier are shown at the bottom.}
	\label{fig:main}
\end{figure*}

\section{Preliminaries}
\subsection{Problem Formulation}

For convenience, we define $\bigl\{X^{(v)}\in \mathbb{R}^{n \times m_{v}}\bigr\}_{v=1}^{l}$ as input data with $l$ views, where $n$ and $m_{v}$ denote the number of samples and dimensionality of $v$-th view. And we let $Y \in \bigl\{0,1\bigr\}^{n \times c}$ represents the label matrix, where $c$ is the number of tags. $Y_{i,:}$ is the label vector of the sample $i$ and $Y_{i,j}=1$ if the sample $i$ belongs to class $j$, otherwise $Y_{i,j}=0$. Additionally, for missing-view and missing-label, we use indicator matrix $\mathcal{W} \in \bigl\{0,1\bigr\}^{n \times l}$ and $\mathcal{G} \in \bigl\{0,1\bigr\}^{n \times c}$ to describe the missing instances distribution, respectively. Specifically, we set $\mathcal{W}_{i,j}=1$ if the instance of $j$-th view corresponding to $i$-th sample is available. Otherwise, we let $\mathcal{W}_{i,j}=0$ for the missing $j$-th view of $i$-th sample and set `NaN' or random noise at the missing position in raw data. Similarly, $\mathcal{G}_{i,j}=1$ means $j$-th tag of $i$-th sample is existed and $\mathcal{G}_{i,j}=0$ for the absence or uncertainty of such a tag. The goal of our DICNet is to learn a reliable classifier to predict multi-class labels for unlabeled test samples by given input, \textit{i.e.}, multi-view data $\bigl\{X^{(v)}\bigr\}_{v=1}^{l}$, mutli-label $Y$, missing-view indicator $\mathcal{W}$, and missing-label indicator $\mathcal{G}$.

\subsection{Relevant IMVL or IMLL Methods}

In the past few years, some methods for IMVL and IMLL have been developed. Yuan et al. proposed a multi-source feature learning method (iMSF), which cleverly divides the incomplete dataset into multiple complete subsets according to missing prior information and obtains a joint representation by imposing the structural sparse regularization constraint \cite{yuan2012multi}. However, iMSF can only handle single-label classification tasks rather than multi-label tasks. Another commendable IMLL algorithm is MvEL-ILD (\textit{i.e.}, multi-view embedding learning for incompletely labeled data), which leverages canonical correlation analysis to map original features to common space and construct similarity matrix to fuse multi-view information \cite{zhang2013multi}. For incomplete semantic labels, MvEL-ILD attempts to construct the graph neighbor constraint based on the correlation of labels' semantic content for consistent prediction results. But MvEL-ILD is only suitable for complete multi-view data, ignoring possible missing-view scenarios. To cope with the DIMVMLC issue, Tan et al. designed a model named iMvWL, which consists of two parts--non-negative MF based IMVL model and label correlation learning based IMLL model \cite{tan2018incomplete}. IMVML bridges the feature space and the semantic space by learning a common representation and imposes low-rank constraint on the label correlation matrix to enhance the predictive power of the model. In addition, Li et al. proposed a non-aligned DIMVMLC model, named NAIML, which introduce a non-aligned constraint to complicate the classification task \cite{li2020concise}. The NAIML is the first to consider the global high-rank property of entire label matrix and the low-rank property of sub-label matrix synchronously.

\section{Methodology}
In this section, we propose a novel deep neural network framework named DICNet for the DIMVMLC task. We will explain our model from the following four aspects: view-specific representation learning, instance-level contrastive learning, incomplete multi-view weighted fusion strategy, and weighted multi-label classification module.

\subsection{View-specific Representation Learning}
It is well known that the raw data contains non-ignorable noise and redundant information, which is not conducive to the learning of semantic content \cite{liu2022localized,wen2022survey}. Therefore, both traditional methods and deep learning methods are devoted to capturing discriminative representation from original feature. Similar to other deep multi-view learning works \cite{wen2020dimc}, we utilize the autoencoder to extract high-level feature instead of focusing on the shallow-level feature like traditional methods. Concretely, our autoencoder is composed of an encoder and a decoder, which are used to extract high-level feature and reconstruct the original data respectively. Each view enjoys its own coder-decoder for capturing the view-specific discriminative information independently. For the $v$-th view, we can define 
$Z^{(v)}=E^{(v)}\Big(X^{(v)},\theta^{(v)}\Big)$ and $\hat{X}^{(v)}=D^{(v)}\Big(Z^{(v)},\psi^{(v)}\Big)$, where $Z^{(v)}\in\mathbb{R}^{n \times d}$ is view-specific high-level feature and $\hat{X}^{(v)}$ denotes the reconstructed feature. $d$ is the pre-defined dimensionality of $Z^{(v)}$. $E^{(v)}$and $D^{(v)}$ represent the encoder and decoder, respectively. $\theta^{(v)}$ and $\psi^{(v)}$ are network parameters corresponding to $E^{(v)}$and $D^{(v)}$. As shown in Fig. \ref{fig:main}, our encoder and decoder can be regarded as two Multilayer Perceptrons (MLPs) with several fully connected (FC) layers. Besides, considering the incomplete multi-view data, following \cite{wen2020dimc}, we introduce the missing-view index matrix $\mathcal{W}$ to avoid the negative effects during the training process, so the weighted reconstruction loss is formulated as:
\begin{equation}
	\scriptsize
	\mathcal{L}_{FR}=\frac{1}{l}\sum_{v=1}^{l}\mathcal{L}_{FR}^{(v)}=\frac{1}{l}\sum_{v=1}^{l}\bigg(\frac{1}{m_{v}}\sum_{i=1}^{n}\Big\|\hat{x}_{i}^{(v)}-x^{(v)}_{i}\Big\|^{2}_{2}\mathcal{W}_{i,v}\bigg),
	\label{eq.lfr}
\end{equation}
where $x^{(v)}_{i}$ and $\hat{x}_{i}^{(v)}$ denote the $i$-th instance of view $v$ and its reconstructed feature. $\mathcal{L}_{FR}^{(v)}$ is the reconstruction loss between input $X^{(v)}$ and output $\hat{X}^{(v)}$, and $\mathcal{L}_{FR}$ represents the mean reconstruction loss of all views.

\begin{figure}[t]

	\centering
	\includegraphics[width=\linewidth]{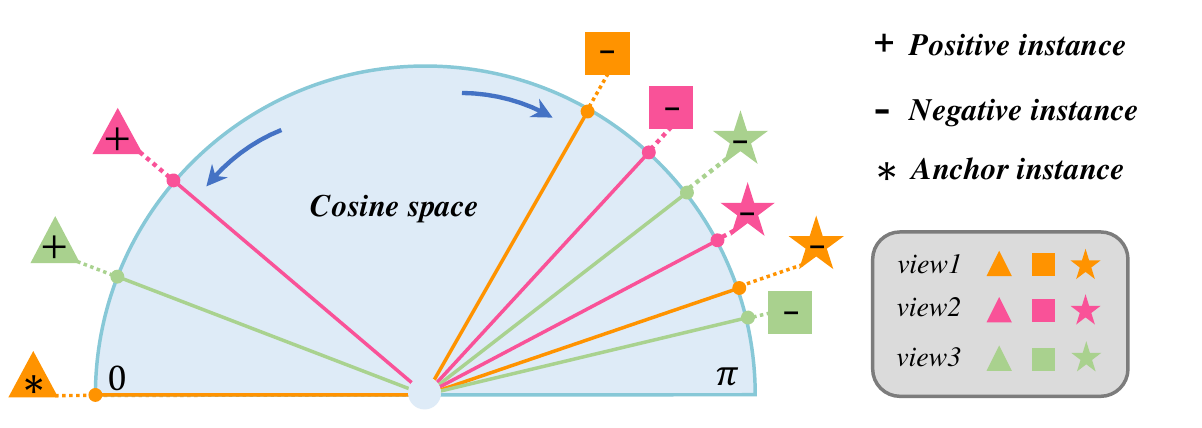}
	\caption{The motivation of instance-level contrastive learning.}
	\label{fig:cont}

\end{figure}
\subsection{Incomplete Instances-level Contrastive Learning}
Through the view-specific representation learning network, we can obtain $l$ high-level representations $\big\{Z^{(v)}\big\}_{v=1}^{l}$ output by the encoders. However, these representations inevitably retain a lot of view-private information since minimizing the reconstruction error will force the encoders to capture the complete information of each view as much as possible, which is obviously not conducive to learning discriminative representations based on the consensus assumption. Considering that the instances, which is belonging to the same sample but from different views, should enjoy similar semantic representation and inspired by \cite{hinton2006fast,chen2020simple,xu2022multi}, we introduce the incomplete instance-level contrastive learning to help the encoders extract more consistent high-level features. 

Specifically, $\big\{Z^{(v)}\big\}_{v=1}^{l}$ learned from the encoders contains $l \times n$ high-level features corresponding to the instances across all views (including missing instances), and we refer to $z^{(v)}_{i} \in \mathbb{R}^d$ as instance of sample \textit{i} in view \textit{v} for convenience. First, we mark all instances in $\big\{Z^{(v)}\big\}_{v=1}^{l}$ with three categories: (1) \textbf{anchor instance} $\mathcal{A}= z^{(v)}_{i}$, (2) \textbf{positive instance} $\mathcal{A}^{+}=z^{(u)}_{i}\Big|_{u\ne v}$ that belongs to the sample \textit{i} but not in view \textit{v}, and (3) \textbf{negative instance} $\mathcal{A}^{-}=z^{(u)}_{j}\Big|_{j\ne i}$ for remainders. It should be noted that the positive and negative instances here are relative to the anchor instance, and each instance can be selected as the anchor instance. Then, we let the anchor instance be paired with another positive or negative instance, so we can get \textit{n-1} positive instance-pairs $\big\{\mathcal{A},\mathcal{A}^{+}\big\}$ and $n \times l-n$ negative instance-pairs $\big\{\mathcal{A},\mathcal{A}^{-}\big\}$, respectively. As shown in Fig. \ref{fig:cont}, it is our motivation that seeks to minimize the distance of available positive instance-pairs and maximize that of available negative instance-pairs in feature space. In our incomplete instance-level contrastive learning, we adopt cosine similarity to measure the distance of instance-pairs \cite{chen2020simple,xu2022multi}:
\begin{equation}
\scriptsize
\mathcal{S}(z_{i}^{(v)} ,z_{j}^{(u)})=\frac{\Big\langle z_{i}^{(v)} \cdot z_{j}^{(u)}\Big\rangle}{\Big\|z_{i}^{(v)}\Big\| \Big\|z_{j}^{(u)}\Big\|},
\end{equation}
where $\langle\cdot\rangle$ denotes the dot product operator, and our optimization goal is $\mathcal{S}(\mathcal{A} ,\mathcal{A}^{+}) \gg \mathcal{S}(\mathcal{A} ,\mathcal{A}^{-})$. It is worth noting that, unlike \cite{xu2022multi}, we introduce the missing-view indicator matrix to exclude unavailable positive instance pairs in the process of calculating contrastive loss. To do this, the contrastive loss between $Z^{(v)}$ and $Z^{(u)}$ is:
\begin{equation}
\scriptsize
l_{IC}^{(vu)}\!=\!-\frac{1}{n}\sum\limits_{i=1}^{n}\mathcal{W}_{i,v}\mathcal{W}_{i,u}\log\frac{\exp({\mathcal{S}\big(z_{i}^{(v)},z_{i}^{(u)}\big)\big/ \tau)}}{\exp({\mathcal{S}\big(z_{i}^{(v)},z_{i}^{(u)}\big)\big/ \tau)}+\mathbb{S}_{neg}},
\label{eq.lvu}
\end{equation}
where $\mathbb{S}_{neg}=\!\sum\limits_{r=u,v}\sum\limits_{j=1, j\ne i}^{n}{\exp({\mathcal{S}\big(z_{i}^{(v)},z_{j}^{(r)}\big)\big/ \tau)}\mathcal{W}_{j,r}}$, and $\tau$ represents the temperature parameter that controls the concentration extent of the distribution \cite{wu2018unsupervised}. Further, the total incomplete instance-level contrastive loss for all view-pairs are as follows:
\begin{equation}
\scriptsize
\mathcal{L}_{IC}=\frac{1}{2}\sum_{v=1}^{l}\sum_{u\ne v}l^{(vu)}_{IC}.
\label{eq.lic}
\end{equation}
As can be seen from Eq.(\ref{eq.lvu}), the contrastive loss for views \textit{v} and \textit{u} is in the form of cross-entropy loss, \textit{i.e.}, minimizing the negative log-likelihood estimator about the similarity distribution of instance pairs. In other words, the contrastive loss of \textit{i}-th sample with respect to \textit{v}-th view and \textit{u}-th view will be computed only if both instances in the positive instance pair $\big\{z^{(v)}_{i},z^{(u)}_{i}\big\}$ are available.

\subsection{Incomplete Multi-view Weighted Fusion}
From the perspective of views, each view inherently enjoys a unique description of the objectives that means the complementarity of view-level information should be exploited to learn a comprehensive sample representation. Indeed, obtaining a view-federated representation is necessary for masses of multi-view learning methods. However, some simple ways, such as concatenating or accumulating the individual representations of all views, are not suitable for incomplete multi-view data due to the possibility of random missing. Therefore, following \cite{wen2020dimc,chen2022adaptively}, a weighted fusion strategy is introduced to combine the multi-view complementary information while avoiding the negative effects of missing views:
\begin{equation}
\scriptsize
h_{i} = \bigg(\sum\limits_{v=1}^{l}z_{i}^{(v)}\mathcal{W}_{i,v}\bigg)\bigg/ \sum\limits_{v=1}^{l}\mathcal{W}_{i,v},
\label{eq.fusion}
\end{equation}
where $z_{i}^{(v)}$ is the view-specific high-level discriminative representation extracted by encoder for $v$-th view of sample $i$. $h_{i}\in \mathbb{R}^{d}$ is the fusion feature for $i$-th sample, \textit{i.e.}, sample-specific representation. Combining all the sample representations from $1$ to $n$, we can obtain the united high-level semantic representation matrix $H=\big\{h_{1}^{T}, h_{2}^{T}, \dots, h_{n}^{T}\big\}$, which is also the input for next classification layer. 



\subsection{Weighted Multi-label Classification Module}
To obtain end-to-end multi-label prediction results, we design a simple classifier to connect the common semantic feature space and label space. Specifically, we expect the classifier to learn a proprietary `template' for each category, which is used to match the sample feature and output the corresponding score. Based on this, we adopt an FC layer with \textit{c} neurons as the main body of the classifier. Besides, a Sigmoid activate function is applied to ensure that the value of prediction is located in the range of $[0,1]$. We formalize the classifier as follows:
\begin{equation}
\setlength{\abovedisplayskip}{6pt}
\setlength{\belowdisplayskip}{6pt}
P=Sigmoid(\mathcal{F}_{c}(H, \omega)),
\end{equation}
where the $\omega $ denotes the learnable parameters of the FC layer, and $P\in \mathbb{R}^{n\times c}$ is our prediction matrix. Finally, following \cite{chen2019learning}, we utilize the binary cross-entropy loss, which is widely used in multi-label classification tasks, as the multi-label classification loss $\mathcal{L}_{MC}$ to evaluate the difference between prediction and ground truth:
\begin{equation}
\scriptsize
\mathcal{L}_{MC}=-\frac{1}{nc}\sum\limits_{i=1}^{n}\sum\limits_{j=1}^{c}\bigg(
\begin{aligned}
&Y_{i,j}\log(P_{i,j})\\
&+(1-Y_{i,j})\log(1-P_{i,j})
\end{aligned}
\bigg)\mathcal{G}_{i,j},
\label{eq.lmc}
\end{equation}
where $Y_{i,j}$ and $P_{i,j}$ represent the real label and prediction, respectively. It is worth noting that we introduce the missing-label indicator $\mathcal{G}$ to filter invalid missing tags, which is similar to the application of missing-view indicator matrix $\mathcal{W}$ in the weighted fusion module.
\begin{algorithm}[t]
	\caption{Semi-supervised training process of DICNet}
	\label{al.1}
	\textbf{Input}: Incomplete multi-view data $\left\{X^{(v)}\right\}_{v=1}^{l}$ with missing-view indicator matrix $\mathcal{W}\in\{0,1\}^{n\times l}$, and corresponding multi-label matrix $Y$ with missing-label indicator matrix $\mathcal{G}\in\{0,1\}^{n\times c}$; Batch size $B$; Hype-parameters $\tau$, $\beta$, and $\lambda$; Training epochs $T$; Stopping threshold $\sigma$.\\
	\textbf{Initialization}: Fill the missing elements of the multi-view data and multi-lable data with `0', and randomly initialize the network weights; Set $\mathcal{L}^{last}=0$; Initialize prediction label $P^{last}$ of $n_t$ test samples.\\
	\textbf{Output}: Parameters of trained model.
	
	\begin{algorithmic}
		\FOR{\textit{k}=1 \textbf{to} \textit{T}}
		\STATE 1.Compute the view-specific representations $\big\{Z^{(v)}\big\}_{v=1}^{l}$ and $\mathcal{L}_{FR}$ using (\ref{eq.lfr}).
		\STATE 2.Compute the instance-level contrastive loss $\mathcal{L}_{IC}$ according to (\ref{eq.lic}).
		\STATE 3.Compute the fusion representation $H$ using (\ref{eq.fusion}).
		\STATE 4.Compute multi-classification loss according to (\ref{eq.lmc}).
		\STATE 5.Compute total loss $\mathcal{L}$ according to (\ref{eq.allloss}) and use the optimizer to update the network parameters batch to batch.
		\STATE 6.Input test samples and obtain prediction $P$. 
		\IF{$|\mathcal{L}^{last}-\mathcal{L}|<\sigma$ \textbf{or} $\frac{1}{n_tc}\sum\limits_{i,j}P_{i,j}\oplus P^{last}_{i,j}<10^{-7}$}
		\STATE Stop training.
		\ELSE
		\STATE Update $\mathcal{L}^{last}$ with $\mathcal{L}$.
		\STATE Update $P^{last}$ with ${P}$.
		\ENDIF
		\ENDFOR
	\end{algorithmic}
	
\end{algorithm}

\subsection{Overall Loss and Complexity Analysis}
The overall training loss of our DICNet can be calculated as:
\begin{equation}
\label{eq.allloss}
\mathcal{L}=\mathcal{L}_{MC}+\beta\mathcal{L}_{IC}+\gamma\mathcal{L}_{FR},
\end{equation}
where $\beta$ and $\gamma$ are penalty parameters for $\mathcal{L}_{IC}$ and $\mathcal{L}_{FR}$, respectively. These losses are calculated only during the training phase, and the weight parameters of the DICNet are updated via back propagation. 
Our DICNet can be trained in a semi-supervised or supervised case, and Algorithm \ref{al.1} shows the training process of DICNet in a semi-supervised manner.

For convenience, we reiterate the relevant symbols, \textit{i.e.}, $n$-the number of samples, $l$-the number of views, $B$-the batch size, $c$-the number of categories, $D$-the maximum number of neurons, $d$-the dimensionality of view-specific representation, and $M$-the maximum dimensionality of raw data. We adopt the min-batch training strategy, so for autoencoders in DNN, the computational complexity is $O(BD^{2}l)$. Besides, the complexity of $\mathcal{L}_{FR}$, $\mathcal{L}_{IC}$, and $\mathcal{L}_{MC}$ is $O(BMl)$, $O(B^{2}dl^{2})$, and $O(Bc)$, respectively. Therefore, the total computational complexity for training of our DICNet is $O\big(\frac{n}{B}(B^{2}dl^{2}+BD^{2}l+BMl+Bc\big)$. We can see that the total complexity increases linearly with the number of samples \textit{n}.

\section{Experiments}
In this section, we introduce our experimental setup and analysis to evaluate our method in detail. And the implementation of our DICNet is based on MindSpore and Pytorch.
\subsection{Experimental Setup}
\textbf{Datasets:} Following \cite{tan2018incomplete,guillaumin2010multimodal,li2020concise}, we select five popular multi-view multi-label datasets in our experiments, \textit{i.e.}, Corel 5k \cite{duygulu2002object}, VOC 2007 \cite{everingham2009pascal}, ESP Game \cite{von2004labeling}, IAPR TC-12 \cite{grubinger2006iapr}, and MIR FLICKR \cite{huiskes2008mir}. For all five multi-view multi-label datasets, we uniformly select six types of features as six views, \textit{i.e.}, GIST, HSV, DenseHue, DenseSift, RGB, and LAB.

\textbf{Double incomplete multi-view and multi-label data preparation:} Following \cite{tan2018incomplete}, we construct double incomplete multi-view multi-label datasets for training and testing based on the above five datasets. For all samples, we first randomly disable $p$\% of instances of every view to build incomplete data (at least one view per sample is available to keep the total number of samples constant). Then, we randomly select $m$ percent of the samples as the training set and the rest as the test set. Finally, we randomly remove $q$\% of positive labels and $q$\% of negative labels. After above three steps, we construct a dataset with $p$\% missing-view rate, $q$\% missing-label rate, and $m$\% training samples. 
\begin{table*}[t!]
\small
\tabcolsep=1.0mm
\begin{center}
\begin{sc}
    \begin{tabular}{ccccccccc}
    \hline
    \hline
    Data & Metric & lrMMC   & MVL-IV   & MvEL-ILD   & iMSF   & iMvWL 	& NAIML	& ours\\
    \midrule	
    \multirow{4}[2]{*}{\begin{turn}{0}Corel 5k\end{turn}}
    &AP	&0.240$\pm$0.002	&0.240$\pm$0.001	&0.204$\pm$0.002	&0.189$\pm$0.002	&0.283$\pm$0.007	&\underline{0.309$\pm$0.004}	&\textbf{{0.381$\pm$0.004}}\\
	&1-HL	&0.954$\pm$0.000	&0.954$\pm$0.000	&0.946$\pm$0.000	&0.943$\pm$0.000	&0.978$\pm$0.000	&\underline{{0.987$\pm$0.000}}	&\textbf{{0.988$\pm$0.000}}\\
	&1-RL	&0.762$\pm$0.002	&0.756$\pm$0.001	&0.638$\pm$0.003	&0.709$\pm$0.005	&0.865$\pm$0.003	&\underline{{0.878$\pm$0.002}}	&\textbf{{0.882$\pm$0.004}}\\
	&AUC	&0.763$\pm$0.002	&0.762$\pm$0.001	&0.715$\pm$0.001	&0.663$\pm$0.005	&0.868$\pm$0.003	&\underline{{0.881$\pm$0.002}}	&\textbf{{0.884$\pm$0.004}}\\
    \midrule
    \multirow{4}[2]{*}{\begin{turn}{0}    VOC 2007\end{turn}}
&AP	&0.425$\pm$0.003	&0.433$\pm$0.002	&0.358$\pm$0.003	&0.325$\pm$0.000	&0.441$\pm$0.017	&\underline{{0.488$\pm$0.003}}	&\textbf{{0.505$\pm$0.012}}\\
&1-HL	&0.882$\pm$0.000	&0.883$\pm$0.000	&0.837$\pm$0.000	&0.836$\pm$0.000	&0.882$\pm$0.004	&\underline{{0.928$\pm$0.001}}	&\textbf{{0.929$\pm$0.001}}\\
&1-RL	&0.698$\pm$0.003	&0.702$\pm$0.001	&0.643$\pm$0.004	&0.568$\pm$0.000	&0.737$\pm$0.009	&\textbf{{0.783$\pm$0.001}}	&\textbf{{0.783$\pm$0.008}}\\
&AUC	&0.728$\pm$0.002	&0.730$\pm$0.001	&0.686$\pm$0.005	&0.620$\pm$0.001	&0.767$\pm$0.012	&\textbf{{0.811$\pm$0.001}}	&\underline{{0.809$\pm$0.006}}\\
    \midrule
    \multirow{4}[2]{*}{\begin{turn}{0}ESP Game\end{turn}}
&AP	&0.188$\pm$0.000	&0.189$\pm$0.000	&0.132$\pm$0.000	&0.108$\pm$0.000	&0.242$\pm$0.003	&\underline{{0.246$\pm$0.002}}	&\textbf{{0.297$\pm$0.002}}\\
&1-HL	&0.970$\pm$0.000	&0.970$\pm$0.000	&0.967$\pm$0.000	&0.964$\pm$0.000	&0.972$\pm$0.000	&\textbf{{0.983$\pm$0.000}}	&\textbf{{0.983$\pm$0.000}}\\
&1-RL	&0.777$\pm$0.001	&0.778$\pm$0.000	&0.683$\pm$0.002	&0.722$\pm$0.002	&0.807$\pm$0.001	&\underline{{0.818$\pm$0.002}}	&\textbf{{0.832$\pm$0.001}}\\
&AUC	&0.783$\pm$0.001	&0.784$\pm$0.000	&0.734$\pm$0.001	&0.674$\pm$0.003	&0.813$\pm$0.002	&\underline{{0.824$\pm$0.002}}	&\textbf{{0.836$\pm$0.001}}\\
    \midrule
    \multirow{4}[2]{*}{\begin{turn}{0}IAPR TC-12\end{turn}}
&AP	&0.197$\pm$0.000	&0.198$\pm$0.000	&0.141$\pm$0.000	&0.101$\pm$0.000	&0.235$\pm$0.004	&\underline{{0.261$\pm$0.001}}	&\textbf{{0.323$\pm$0.001}}\\
&1-HL	&0.967$\pm$0.000	&0.967$\pm$0.000	&0.963$\pm$0.000	&0.960$\pm$0.000	&0.969$\pm$0.000	&\textbf{{0.981$\pm$0.000}}	&\textbf{{0.981$\pm$0.000}}\\
&1-RL	&0.801$\pm$0.000	&0.799$\pm$0.001	&0.725$\pm$0.001	&0.631$\pm$0.000	&0.833$\pm$0.003	&\underline{{0.848$\pm$0.001}}	&\textbf{{0.873$\pm$0.001}}\\
&AUC	&0.805$\pm$0.000	&0.804$\pm$0.001	&0.746$\pm$0.001	&0.665$\pm$0.001	&0.836$\pm$0.002	&\underline{{0.850$\pm$0.001}}	&\textbf{{0.874$\pm$0.001}}\\
    \midrule
    \multirow{4}[2]{*}{\begin{turn}{0}MIR FLICKR\end{turn}}
    &AP	&0.441$\pm$0.001	&0.449$\pm$0.001	&0.375$\pm$0.000	&0.323$\pm$0.000	&0.495$\pm$0.012	&\underline{{0.551$\pm$0.002}}	&\textbf{{0.589$\pm$0.005}}\\
	&1-HL	&0.839$\pm$0.000	&0.839$\pm$0.000	&0.778$\pm$0.000	&0.775$\pm$0.000	&0.840$\pm$0.003	&\underline{{0.882$\pm$0.001}}	&\textbf{{0.888$\pm$0.002}}\\
	&1-RL	&0.802$\pm$0.001	&0.808$\pm$0.001	&0.771$\pm$0.001	&0.641$\pm$0.001	&0.806$\pm$0.011	&\underline{{0.844$\pm$0.001}}	&\textbf{{0.863$\pm$0.004}}\\
	&AUC	&0.806$\pm$0.001	&0.807$\pm$0.000	&0.761$\pm$0.000	&0.715$\pm$0.001	&0.794$\pm$0.015	&\underline{{0.837$\pm$0.001}}	&\textbf{{0.849$\pm$0.004}}\\
\hline
\hline
\end{tabular}
\end{sc}
\end{center}
\caption{Experimental results of different methods on the five datasets with 50\% missing instances, 70\% training samples, and 50\% missing labels for training samples. The 1st/2nd best resluts are marked in bold/underline.}

\label{table1}
\end{table*}

\textbf{Compared with related methods}: In our experiments, we select six state-of-the-art methods to compare with our DICNet. Four of them are introduced in the preliminaries, \textit{i.e.}, MvEL-ILD, iMSF, iMvWL, and NAIML. In addition, we briefly introduce the other two comparison methods: (1) lrMMC \cite{liu2015low}, a complete MF based MvMLC method, attempt to preserve the low-rank property of original features. (2) MVL-IV \cite{xu2015multi} is an IMVL model based on the missing-view recovery strategy. Notably, only iMVWL and NAIML can handle both incomplete views and incomplete tags, so we have to make some adjustments to the other four methods as \cite{tan2018incomplete} did: For MvEL-ILD and lrMMC, we fill missing-view with average value; For iMSF and MVL-IV, corresponding missing tags are set as negative tags. In addition, for a fair comparison, optimal parameters for the six approaches are selected as mentioned in their papers, and ten replicates are conducted to reduce the randomness of results.

\textbf{Evaluation metrics}: Similar to \cite{tan2018incomplete} and \cite{li2020concise}, four popular metrics commonly used in the multi-label learning field are adopted to evaluate these approaches.\textit{ i.e.}, Ranking Loss (RL), Average Precision (AP), Hamming Loss (HL), and adapted area under curve (AUC) \cite{bucak2011multi,zhang2013review}. In particular, to facilitate the observation and comparison of the performance of different methods, we show the value of 1-RL and 1-HL instead of RL and HL in our report. Thus, a more intuitive criterion for comparison is: higher values of the four metrics mean better performance.
\begin{figure}[t!]
		\centering
		\subfloat[missing-view study]{
			\label{fig:rates-a}
			\includegraphics[width=0.48\linewidth]{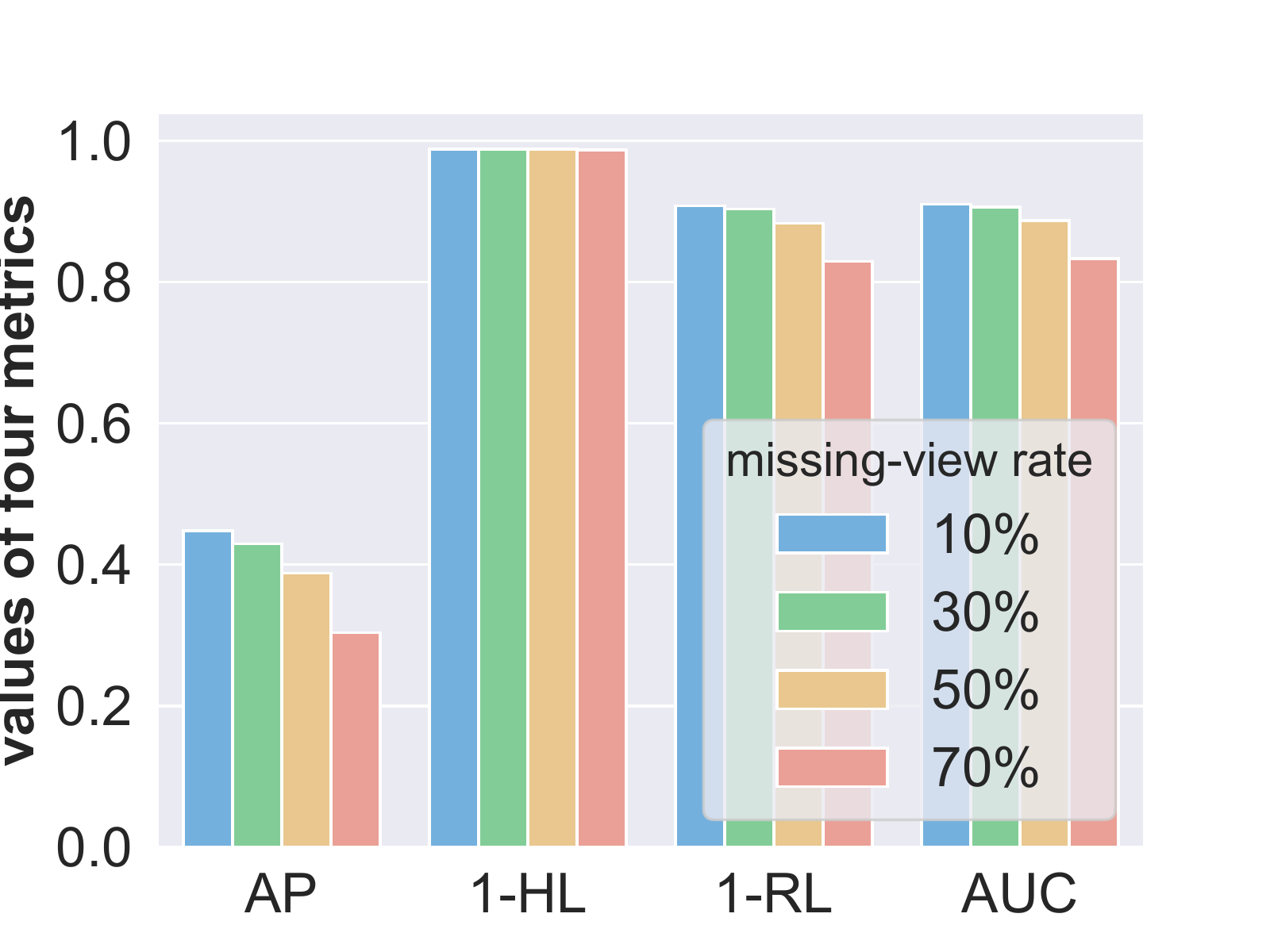}
		}
		\subfloat[missing-label study]{
			\label{fig:rates-b}
			\includegraphics[width=0.48\linewidth]{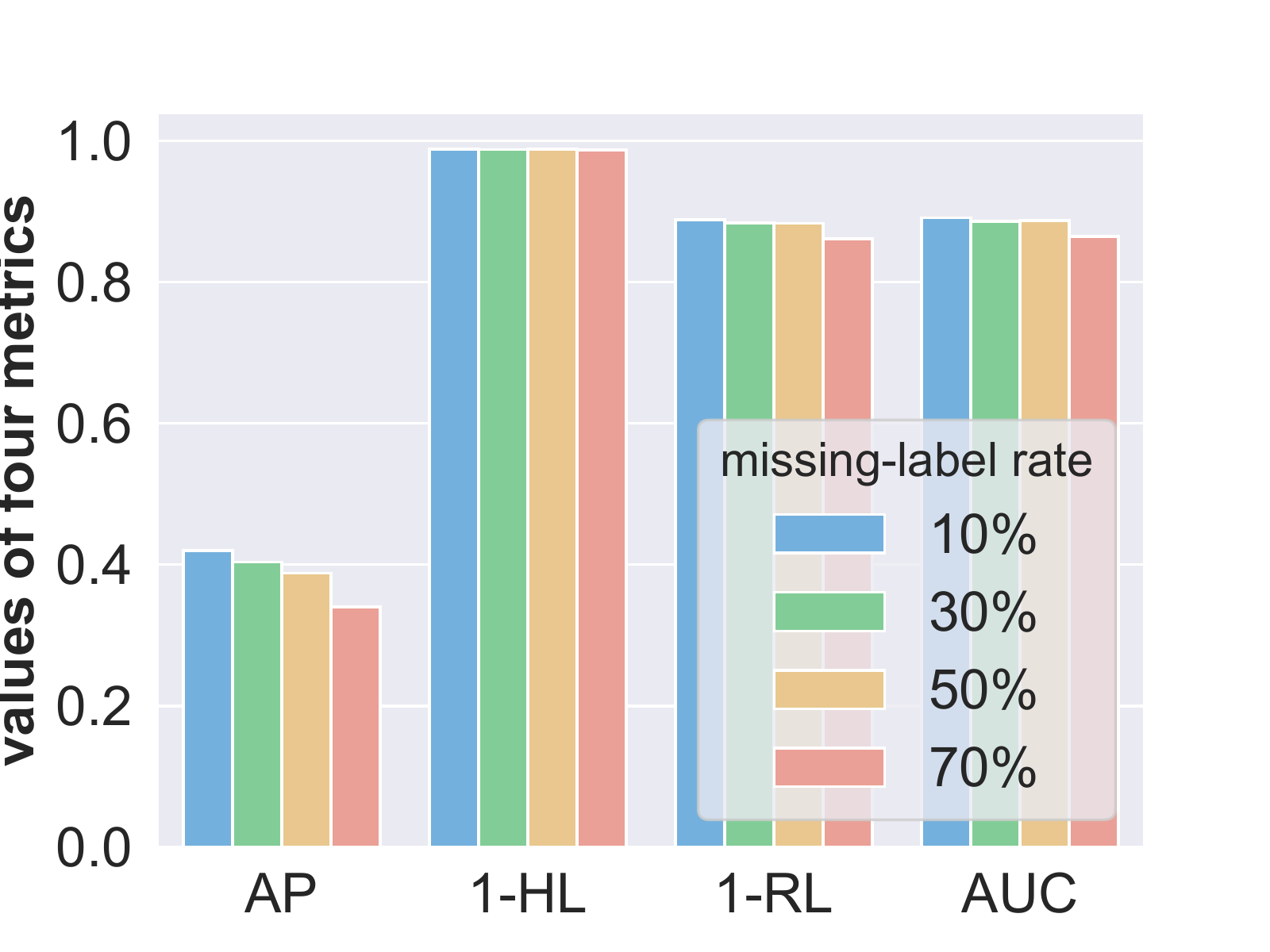}
		}

		\caption{The results on Corel5k dataset with (a) different missing-view rates and a 50\% missing-label rate and (b) a 50\% missing-view rate and different missing-label rates.}
		\label{fig:rates}
		
\end{figure}

\subsection{Experimental Results and Analysis}
The statistical results of repeated experiments of seven methods on five aforementioned databases with 50\% missing-view rate, 50\% missing-label rate, and 70\% training samples are listed in Table \ref{table1}. And the results of the comparison methods are quoted from \cite{li2020concise} and \cite{tan2018incomplete}. Values in parentheses represent the standard deviation. From the Table \ref{table1}, it is easy to obtain the following information:
\begin{itemize}
\item Compared with the first four methods, which are not designed for DIMVMLC tasks, the iMvWL, NAIML, and our DICNet enjoy obvious performance advantages on all five datasets. For instance, the values about AP of iMvWL and DICNet exceed lrMMC by 7 and 14 percentage points on the Corel5k dataset, respectively. We have reason to believe that the weighted fusion strategy based on prior missing information plays a positive role in reducing the negative effects of missing views and labels. It is this comprehensive consideration that helps the model adapt to DIMVMLC tasks.

\item In comparison with the other six methods, our proposed DICNet has a bright performance, which is top in almost all metrics. In particular, on the most representative AP value, our DICNet is about 8 percentage points higher than the second-best NAIML on the corel5k database. Even on large-scale MIR FLICKR dataset with 25, 000 samples, its lead remained significant. These results illustrate that DNN is able to extract high-level discriminative features more effectively than traditional methods.
\end{itemize}

To further investigate the impact of different missing-view and missing-label ratios on classification performance, we conduct our DICNet on the Corel5k dataset and report the results in Fig. \ref{fig:rates}. Specifically, we fix one incomplete ratio at 50\%, and then alter another incomplete ratio to 0, 30\%, 50\%, and 70\%, respectively to observe the variation trends of four metrics. From Fig. \ref{fig:rates}, we can distinctly see that: (1) As the incompleteness rate of views or labels increases, the values of four metrics, especially the AP value, gradually decrease. (2) At the same miss rate, partial views have a greater impact on performance than partial tags to some extent. These intuitive phenomena once again verify the harmfulness of missing views and missing labels. Meanwhile, it can be inferred that our model is more dependent on the extraction and learning of high-level features than the supervision information in labels, which is still an open question.
\begin{figure}[t!]
		
		\centering
		\subfloat[Corel5k]{
			\label{fig:param1}
			\includegraphics[width=0.49\linewidth,height=3cm]{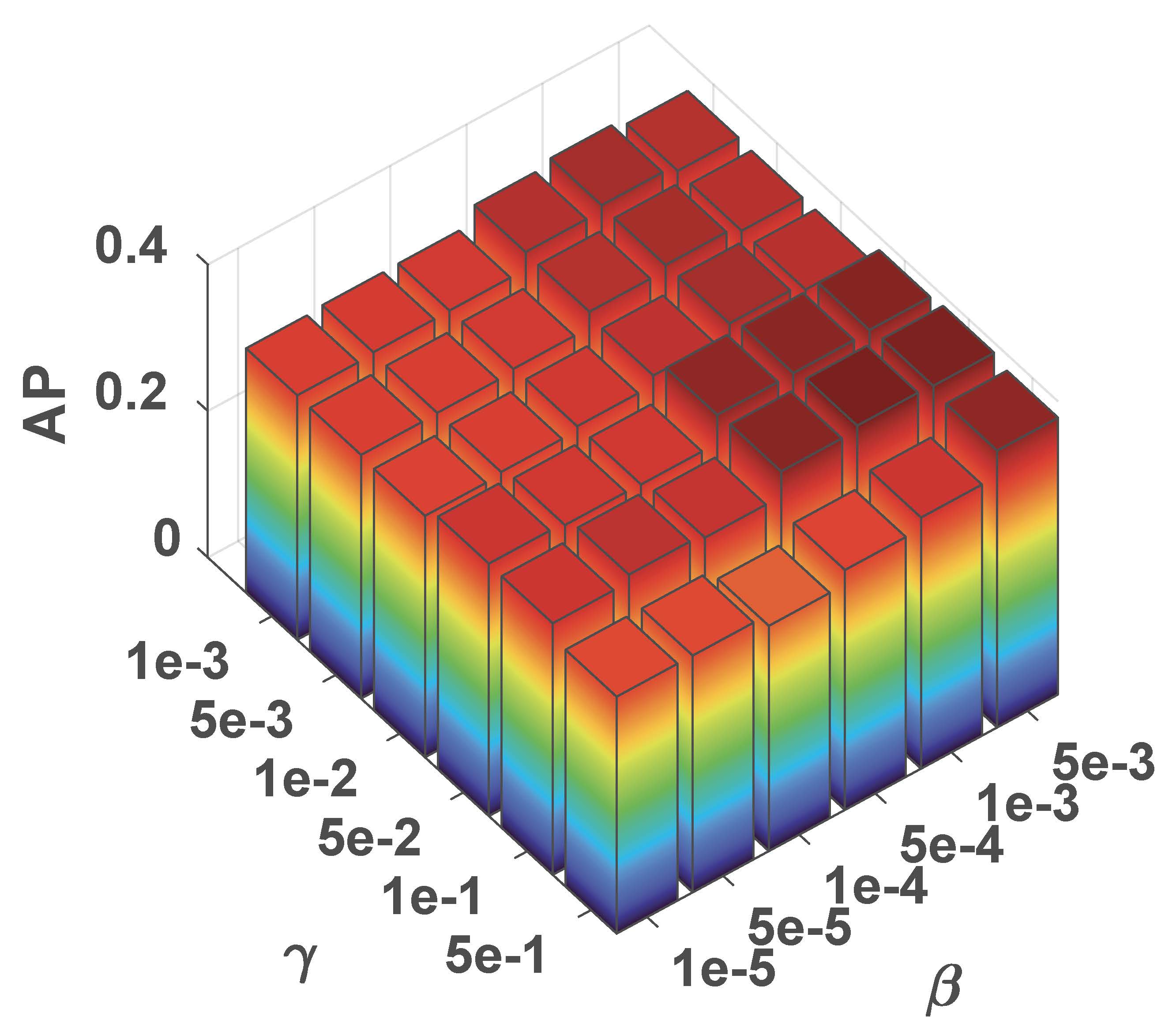}
		}
		\subfloat[VOC2007]{
			\label{fig:param2}
			\includegraphics[width=0.49\linewidth,height=3cm]{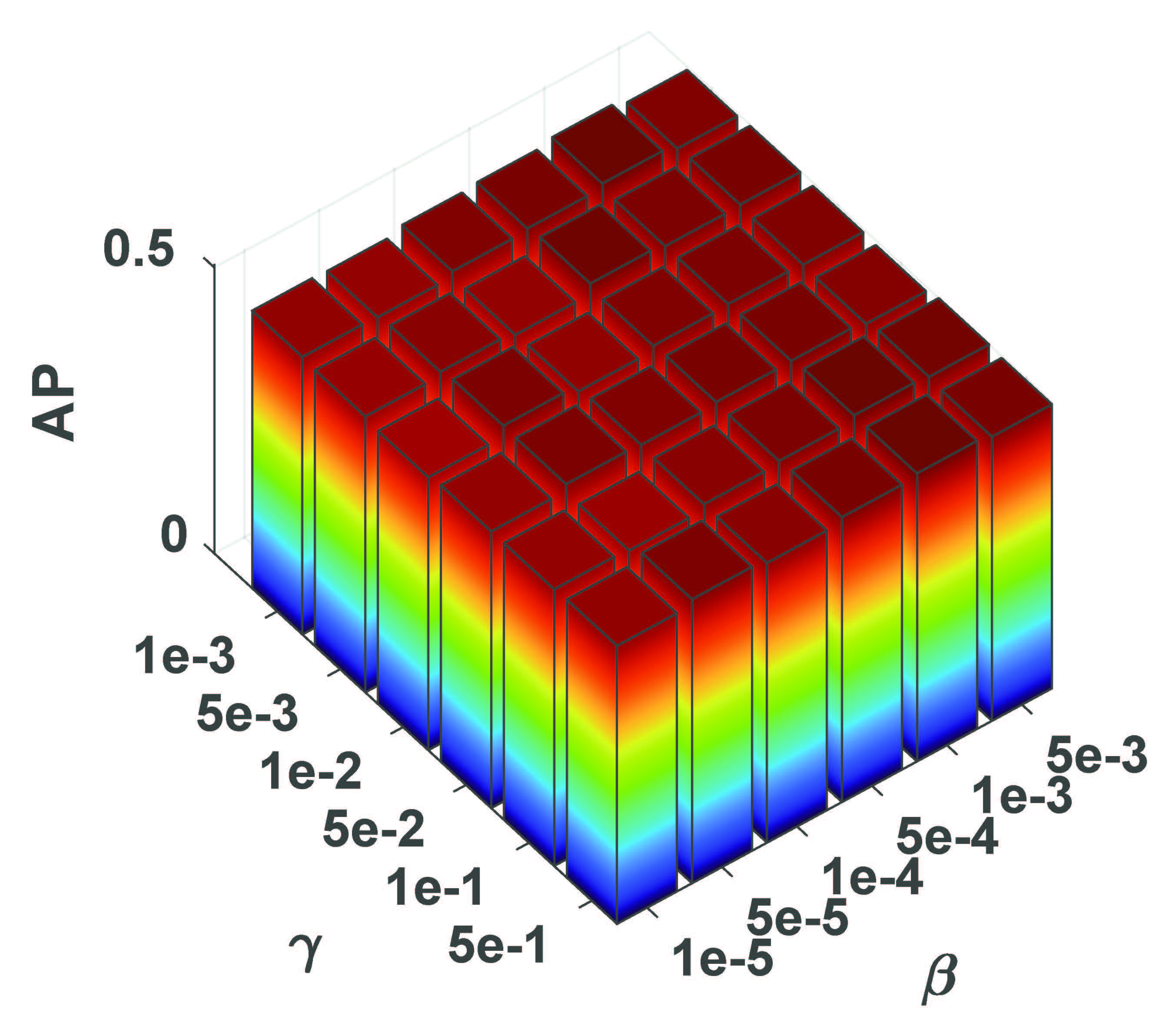}
		}
		\quad
		\subfloat[Corel5k]{
			\label{fig:param3}
			\includegraphics[width=0.48\linewidth]{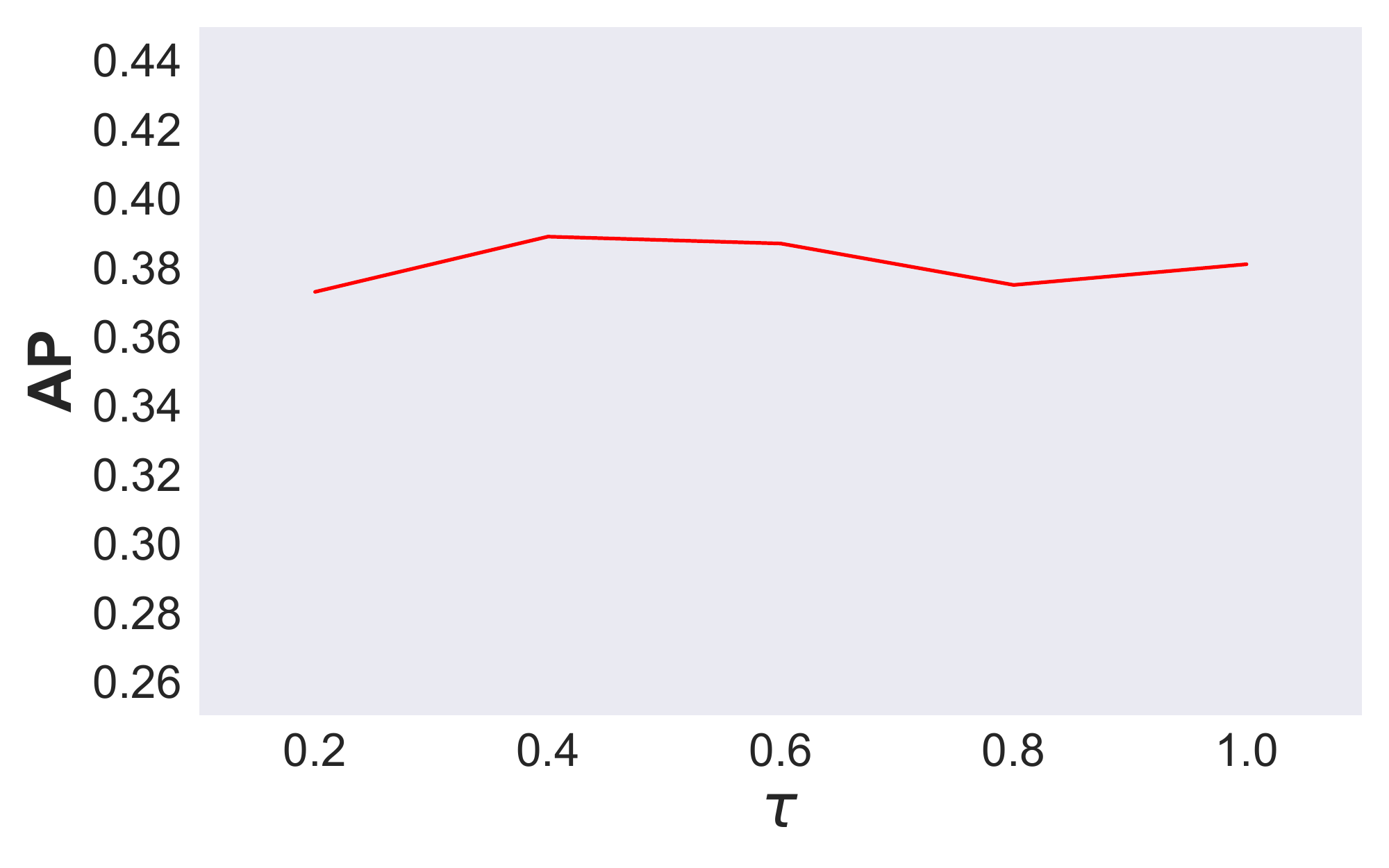}
		}
		\subfloat[VOC2007]{
			\label{fig:param4}
			\includegraphics[width=0.48\linewidth]{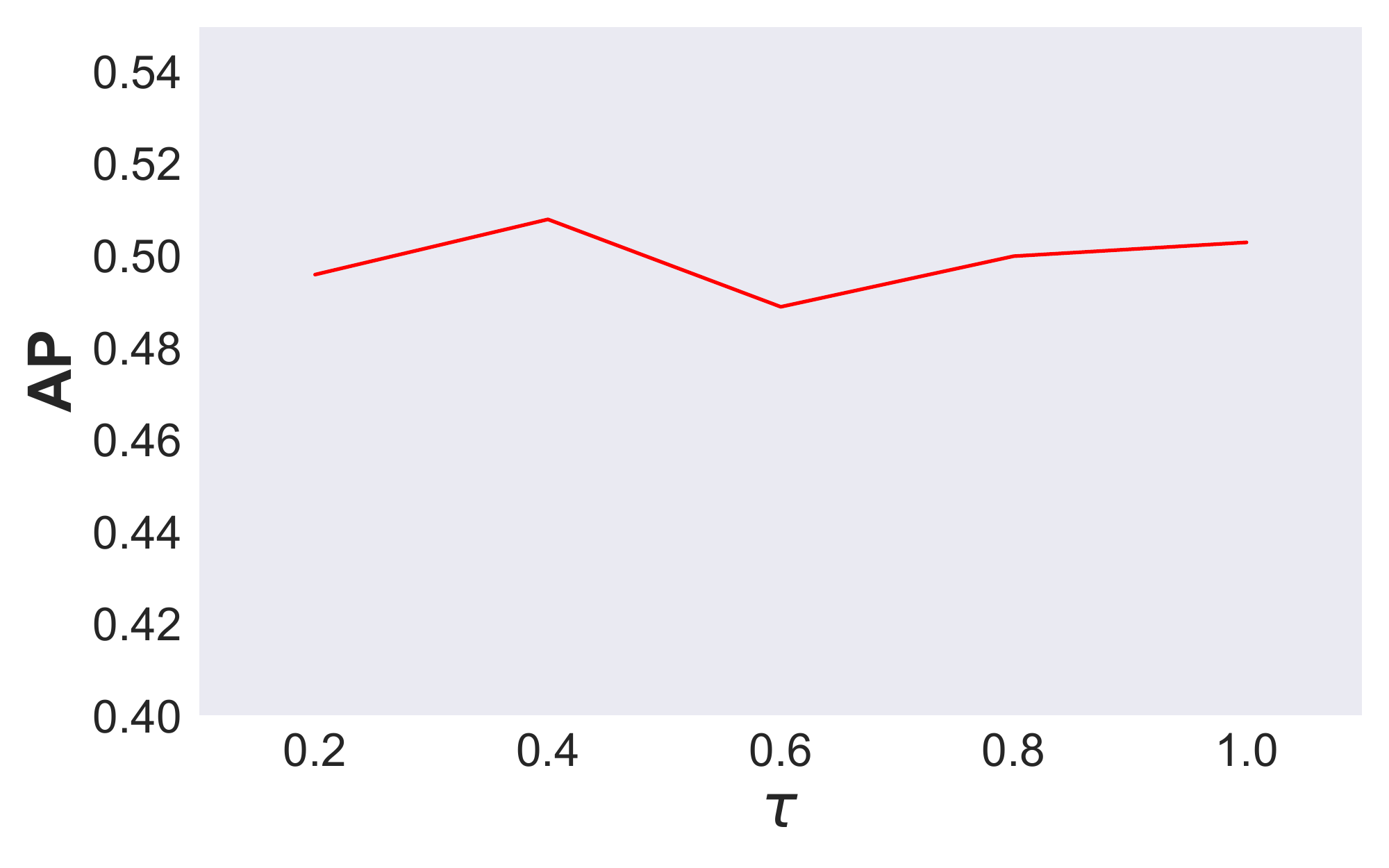}
		}
		 
		\caption{The AP value versus $\beta$ and $\gamma$ on the (a) Corel5k dataset and (b) VOC2007 dataset; the AP value versus $\tau$ on the (c) Corel5k dataset and (d) VOC2007 dataset. All datasets are with 50\% available instances, 50\% missing labels, and 70\% training samples.}
		\label{fig:params}
		
\end{figure}

\subsection{Hyper-parameter Analysis and Ablation Study}
In our DICNet, there are three hyper-parameters, \textit{i.e.}, $\beta$, $\gamma$, and $\tau$ that need to be set before training. In order to study the sensitivity of our DICNet to the three hyper-parameters, we experiment on the corel5k dataset and pascal07 dataset with 50\% available instances, 50\% missing labels, and 70\% training samples. Fig. \ref{fig:param1} and Fig. \ref{fig:param2} show the AP value versus hyper-parameters $\beta$ and $\gamma$. Fig. \ref{fig:param3} and Fig. \ref{fig:param4} plot the curves of AP \textit{w.r.t} the selection of $\tau$ respectively. Irrelevant hyper-parameters are fixed to guarantee the validity of all experiments. Obviously, when the hyper-parameters $\beta$ and $\gamma$ are correspondingly selected from the ranges of [5e-4, 5e-3] and [5e-2, 1e-1] for the corel5k dataset and [5e-3, 1e-5] and [1e-3, 5e-1] for the pascal07 dataset, our DICNet can achieve relatively stable and satisfactory performance. As for temperature parameter $\tau$, it seems to have an inappreciable impact on performance, so we set it to 0.5 for all datasets. 

To verify the effectiveness of various parts of our method, we perform ablation experiments on the corel5k and pascal07 datasets with 50\% instances, 50\% missing labels, and 70\% training samples. First, we select the multi-label classification loss $\mathcal{L}_{MC}$, which is essential for supervised or semi-supervised classification tasks, as our benchmark, and the benchmark shows comparable performance compared to NAIML. Then, we superimpose $\mathcal{L}_{FR}$ and $\mathcal{L}_{IC}$, step by step on the benchmark. Finally, from Table \ref{table3}, we can find that: (1) with the superposition of each loss component, the performance metrics increase significantly; (2) the biggest improvement comes from the addition of $\mathcal{L}_{IC}$. These phenomena illustrate that all parts of our model have gains for high-performance multi-label classification, especially for our incomplete instance-level contrastive learning.

\begin{table}[t]

\begin{center}
\small
	\begin{tabular}{ccc|cc|cc}
	    \hline
	    \toprule
	
				\multirow{2}{*}{$\mathcal{L}_{MC}$} & \multirow{2}{*}{$\mathcal{L}_{FR}$} & \multirow{2}{*}{$\mathcal{L}_{IC}$}	& \multicolumn{2}{c|}{Corel5k} & \multicolumn{2}{c}{VOC2007} \\ 
		        ~ & ~ & ~ & AP & AUC & AP & AUC \\
		        \midrule	\Checkmark & ~ & ~ &	0.336 & 0.858 & 0.484 & 0.788 \\ 
		        \cline{4-7}	\Checkmark & \Checkmark & ~ & 0.352 & 0.872 & 0.492 & 0.802 \\ 
		        \cline{4-7}	\Checkmark & ~ & \Checkmark  & 0.368 & 0.876 & 0.504 & 0.809 \\ 
		        \cline{4-7}	\Checkmark & \Checkmark & \Checkmark &  0.381 & 0.884 & 0.505 & 0.809 \\ 

	    \bottomrule
	    \hline
		\end{tabular}

\end{center}
\caption{Ablation experimental results of our DICNet on the Corel5k and VOC2007 datasets with 50\% instances, 50\% missing labels, and 70\% training samples.}
\label{table3}%
\end{table}

\section{Conclusion}
In this paper, we propose an ingenious neural network model for the DIMVMLC tasks. Most notably, we design the instance-level contrastive loss to guide the autoencoders to learn the cross-view high-level representation based on the consensus assumption. Moreover, a partial multi-view weighted fusion strategy is developed to exploit complementary information and enhance the discriminative ability. Throughout the learning model, we utilize the view and label missing indicator matrices to cleverly avoid the deleterious effects of incompleteness. Finally, complete and convincing experimental results confirm that our method is reliable and advanced compared to other state-of-the-art methods.

\section*{Acknowledgments}
This work is supported by Shenzhen Science and Technology Program under Grant RCBS20210609103709020, GJHZ20210705141812038, National Natural Science Foundation of China under Grant 62006059, and CAAI-Huawei MindSpore Open Fund under Grant CAAIXSJLJJ-2022-011C.

\bibliography{aaai23}

\bigskip

\end{document}